# Information-theoretic Semi-supervised Metric Learning via Entropy Regularization

**Gang Niu**  GANG@SG.CS.TITECH.AC.JP
Department of Computer Science, Tokyo Institute of Technology, Tokyo 152-8552, Japan

**Bo Dai**  BOHR.DAI@GMAIL.COM
Department of Computer Science, Purdue University, West Lafayette, IN 47907, USA

**Makoto Yamada**  YAMADA@SG.CS.TITECH.AC.JP
**Masashi Sugiyama**  SUGI@CS.TITECH.AC.JP
Department of Computer Science, Tokyo Institute of Technology, Tokyo 152-8552, Japan

## Abstract

We propose a general information-theoretic approach called SERAPH *(SEmi-supervised metRic leArning Paradigm with Hyper-sparsity)* for metric learning that does not rely upon the manifold assumption. Given the probability parameterized by a Mahalanobis distance, we maximize the entropy of that probability on labeled data and minimize it on unlabeled data following *entropy regularization*, which allows the supervised and unsupervised parts to be integrated in a natural and meaningful way. Furthermore, SERAPH is regularized by encouraging a low-rank projection induced from the metric. The optimization of SERAPH is solved efficiently and stably by an EM-like scheme with the analytical E-Step and convex M-Step. Experiments demonstrate that SERAPH compares favorably with many well-known global and local metric learning methods.

## 1. Introduction

A good metric for input data is a key factor for many machine learning algorithms. Classical metric learning methods fall into three types: (a) Supervised type requiring class labels (e.g., Sugiyama, 2007); (b) Supervised type requiring weak labels, i.e., $\{\pm 1\}$-valued labels that indicate the similarity/dissimilarity of data pairs (e.g., Weinberger et al., 2005; Davis et al., 2007); (c) Unsupervised type requiring no label information (e.g., Belkin & Niyogi, 2001). Types (a) and (b) have a strict limitation for real-world applica-tions since they need lots of labels. Based on the belief that preserving the geometric structure in an unsupervised manner can be better than relying on the limited labels, semi-supervised metric learning has emerged. To the best of our knowledge, all semi-supervised extensions employ *off-the-shelf* techniques in type (c) such as principal component analysis (Yang et al., 2006; Sugiyama et al., 2010) or manifold embedding (Hoi et al., 2008; Baghshah & Shouraki, 2009; Liu et al., 2010). They can be regarded as propagating labels along an assistant metric by some unsupervised techniques and learning a target metric implicitly in a supervised manner.

However, the target and assistant metrics assume different forms, one Mahalanobis distance defined over a Euclidean space and one geodesic distance over a curved space or a Riemannian manifold. The two metrics also share slightly different goals: the target metric tries to learn a metric so that data in the same class are close and data from different classes are far apart (e.g., Fisher discriminant analysis[1] (Fisher, 1936)), and the assistant one tries to identify and preserve the intrinsic geometric structure (e.g., Laplacian eigenmaps (Belkin & Niyogi, 2001)). Simply putting them together works in practice, but the paradigm is conceptually neither natural nor unified.

In this paper, we propose a semi-supervised metric learning approach SERAPH (SEmi-supervised metRic leArning Paradigm with Hyper-sparsity) as an *information-theoretic alternative* to the manifold-based methods. Our idea is to optimize a metric by optimizing a conditional probability parameterized by that metric. Following *entropy regularization* (Grandvalet & Bengio, 2004), we maximize the entropy of that probability on labeled data, and minimize it on unlabeled data, which can achieve the sparsity of the posterior distribution (Graça et al., 2009), i.e., the low uncertainty/entropy of unobserved weak labels. Furthermore, we employ *mixed-norm regularization* (Ying et al., 2009) to encourage the sparsity of the projection matrix, i.e., the low rank of the projection matrix induced from the metric. Unifying the posterior sparsity and the projection sparsity brings us to the *hyper-sparsity*. Thanks to this property, the metric learned by SERAPH possesses high discriminability even under a noisy environment.

---
[1] Note that learning a metric is equivalent to learning a projection in the scenario of dimensionality reduction.

Our contributions can be summarized as follows. First, we formulate the supervised metric learning problem as an instance of the generalized maximum entropy distribution estimation (Dudík & Schapire, 2006). Second, we propose a semi-supervised extension of the above estimation following entropy regularization (Grandvalet & Bengio, 2004). Notice that our extension is compatible with the manifold-based extension, which means that SERAPH could adopt an additional manifold regularization term.

## 2. Proposed Approach

In this section, we first formulate the model of SERAPH and then develop the EM-like algorithm to solve the model.

### 2.1. Notations

Suppose we have a training set $\mathcal{X} = \{x_i \mid x_i \in \mathbb{R}^m\}_{i=1}^n$ that contains $n$ points each with $m$ features. Let the sets of similar and dissimilar data pairs be

$$\mathcal{S} = \{(x_i, x_j) \mid x_i \text{ and } x_j \text{ are similar}\},$$
$$\mathcal{D} = \{(x_i, x_j) \mid x_i \text{ and } x_j \text{ are dissimilar}\}.$$

With some abuse of terminology, we refer to $\mathcal{S} \cup \mathcal{D}$ as the labeled data and

$$\mathcal{U} = \{(x_i, x_j) \mid i \neq j, (x_i, x_j) \notin \mathcal{S} \cup \mathcal{D}\}$$

as the unlabeled data. A weak label $y_{i,j} = 1$ is assigned to $(x_i, x_j) \in \mathcal{S}$, or $y_{i,j} = -1$ to $(x_i, x_j) \in \mathcal{D}$. We abbreviate $\sum_{(x_i,x_j)\in\mathcal{S}\cup\mathcal{D}}$, $\sum_{(x_i,x_j)\in\mathcal{U}}$ and $\sum_{y\in\{1,-1\}}$ as $\sum_{\mathcal{S}\cup\mathcal{D}}$, $\sum_\mathcal{U}$ and $\sum_y$. Consider learning a Mahalanobis distance metric for $x, x' \in \mathbb{R}^m$ of the form

$$d(x, x') = \|x - x'\|_A = \sqrt{(x - x')^\top A(x - x')},$$

where $^\top$ is the transpose operator and $A \in \mathbb{R}^{m \times m}$ is a symmetric and positive semi-definite matrix to be learned[2]. The probability $p^A(y \mid x, x')$ of labeling $(x, x') \in \mathbb{R}^m \times \mathbb{R}^m$ with $y = \pm 1$ is parameterized by the matrix $A$. When applying $p^A(y \mid x, x')$ to $(x_i, x_j)$, it is abbreviated as $p_{i,j}^A(y)$.

---
[2] In this paper, $A$ is always assumed symmetric positive semi-definite and will not be explicitly written for brevity.

### 2.2. Basic model

To begin with, we derive a probabilistic model to investigate the conditional probability of $y = \pm 1$ given $(x, x') \in \mathbb{R}^m \times \mathbb{R}^m$. We resort to a parametric form of $p^A(y \mid x, x')$, and will focus on it for the out-of-sample ability.

The *maximum entropy principle* (Jaynes, 1957) suggests that we should choose the probability distribution with the maximum entropy out of all distributions that match the data moments. Let[3]

$$H(p_{i,j}^A) = -\sum_y p_{i,j}^A(y) \ln p_{i,j}^A(y)$$

be the entropy of the conditional probability $p_{i,j}^A(y)$, and

$$f(x, x', y; A) : \mathbb{R}^m \times \mathbb{R}^m \times \{+1, -1\} \mapsto \mathbb{R}$$

be a feature function that is convex with respect to $A$. The constrained optimization problem is

$$\max_{A, p_{i,j}^A, \xi} \sum_{\mathcal{S}\cup\mathcal{D}} H(p_{i,j}^A) - \frac{1}{2\gamma}\xi^2$$
$$\text{s.t.} \left|\sum_{\mathcal{S}\cup\mathcal{D}} \mathbb{E}_{p_{i,j}^A}[f(x_i, x_j, y; A)] - \sum_{\mathcal{S}\cup\mathcal{D}} f(x_i, x_j, y_{i,j}; A)\right| \leq \xi, \quad (1)$$

where $\xi$ is a slack variable and $\gamma > 0$ is a regularization parameter. The penalty presumes the Gaussian prior of the expected data moments from the empirical data moments, which is essentially consistent in spirit with the *generalized maximum entropy principle* (Dudík & Schapire, 2006) (see Appendix B.1).

**Theorem 1.** *The primal solution $p^{*A}$ is given in terms of the dual solution $(A^*, \kappa^*)$ by*

$$p^{*A}(y \mid x, x') = \frac{\exp(\kappa^* f(x, x', y; A^*))}{Z(x, x'; A^*, \kappa^*)}, \quad (2)$$

*where $Z(x, x'; A, \kappa) = \sum_{y'} \exp(\kappa f(x, x', y'; A))$, and $(A^*, \kappa^*)$ can be obtained by solving the dual problem*

$$\min_{A, \kappa} \sum_{\mathcal{S}\cup\mathcal{D}} \ln Z(x_i, x_j; A, \kappa) - \sum_{\mathcal{S}\cup\mathcal{D}} \kappa f(x_i, x_j, y_{i,j}; A) + \frac{\gamma}{2}\kappa^2. \quad (3)$$

*Define the regularized log-likelihood function on labeled data (i.e., on observed weak labels) as*

$$\mathcal{L}_1(A, \kappa) = \sum_{\mathcal{S}\cup\mathcal{D}} \ln p_{i,j}^A(y_{i,j}) - \frac{\gamma}{2}\kappa^2.$$

*Then, for supervised metric learning, the regularized maximum log-likelihood estimation and the generalized maximum entropy estimation are equivalent.*[4]

---
[3] Throughout this paper, we adopt that $0 \ln 0 = 0$.
[4] The proofs of all theorems are in Appendix A.

When considering $f(x, x', y; A)$ that should take moments about the metric information into account, we propose

$$f(x, x', y; A, \eta) = \frac{y}{2}(\|x - x'\|_A^2 - \eta), \quad (4)$$

where $\eta > 0$ is a hyperparameter used as the threshold to separate the sets $\mathcal{S}$ and $\mathcal{D}$ under the target metric $d(x, x')$. Now the probabilistic model (2) becomes

$$p^A(y \mid x, x') = \frac{1}{1 + \exp(-\kappa y(\|x - x'\|_A^2 - \eta))}. \quad (5)$$

For the optimal solution $(p^{*A}, A^*, \kappa^*)$, we hope for

$$p^{*A}(y_{i,j} \mid x_i, x_j) > 1/2, \ y_{i,j}(\|x_i - x_j\|_{A^*}^2 - \eta) < 0,$$

so there must be $\kappa^* < 0$.

Although we use Eq.(4) as our feature function, other options are available. Please see Appendix C.1 for details.

### 2.3. Regularization

In this subsection, we extend $\mathcal{L}_1(A, \kappa)$ by entropy regularization to semi-supervised learning. Moreover, we regularize our objective by trace-norm regularization.

Our unsupervised part does not rely upon the manifold assumption and is not in the paradigm of smoothing the projected training data. In order to be integrated with the supervised part more naturally in philosophy, we follow the *minimum entropy principle* (Grandvalet & Bengio, 2004), and hence $p_{i,j}^A$ should have low entropy or uncertainty for $(x_i, x_j) \in \mathcal{U}$. Roughly speaking, the resultant discriminative models prefer peaked distributions on unlabeled data, which carries out a probabilistic *low-density separation*. Subsequently, according to Grandvalet & Bengio (2004), our optimization becomes

$$\max_{A,\kappa} \ \mathcal{L}_2(A, \kappa) = \sum_{\mathcal{S} \cup \mathcal{D}} \ln p_{i,j}^A(y_{i,j}) - \frac{\gamma}{2}\kappa^2$$
$$+ \mu \sum_{\mathcal{U}} \sum_y p_{i,j}^A(y) \ln p_{i,j}^A(y),$$

where $\mu \geq 0$ is a regularization parameter.

In addition, we hope for the dimensionality reduction ability by encouraging a low-rank projection induced from $A$. This is helpful in dealing with corrupted data or data distributed intrinsically in a low-dimensional subspace. It is known that the trace is a convex relaxation of the rank for a matrix, so we revise our optimization problem into

$$\max_{A,\kappa} \ \mathcal{L}(A, \kappa) = \sum_{\mathcal{S} \cup \mathcal{D}} \ln p_{i,j}^A(y_{i,j}) - \frac{\gamma}{2}\kappa^2$$
$$+ \mu \sum_{\mathcal{U}} \sum_y p_{i,j}^A(y) \ln p_{i,j}^A(y) - \lambda \operatorname{tr}(A), \quad (6)$$

where $\operatorname{tr}(A)$ is the trace of $A$, and $\lambda \geq 0$ is a regularization parameter.

Optimization (6) is the final model of SERAPH, and we say that it is equipped with the hyper-sparsity when both $\mu$ and $\lambda$ are positive. SERAPH possesses standard kernel and manifold extensions. For more information, please refer to Appendix C.2 and C.3.

### 2.4. Algorithm

From now on we will simplify the model (6) and derive a practical algorithm. First, we eliminate $\kappa$ from (6), thanks to the fact that we use a simple feature function (4) in (1).

**Theorem 2.** *Define the simplified optimization problem as*[5]

$$\max_A \ \hat{\mathcal{L}}(A) = \sum_{\mathcal{S} \cup \mathcal{D}} \ln \hat{p}_{i,j}^A(y_{i,j})$$
$$+ \mu \sum_{\mathcal{U}} \sum_y \hat{p}_{i,j}^A(y) \ln \hat{p}_{i,j}^A(y) - \hat{\lambda} \operatorname{tr}(A), \quad (7)$$

*where the simplified probabilistic model is*

$$\hat{p}^A(y \mid x, x') = \frac{1}{1 + \exp(y(\|x - x'\|_A^2 - \hat{\eta}))}. \quad (8)$$

*Let $\hat{A}$ and $(A^*, \kappa^*)$ be the optimal solutions to (7) and (6), respectively. Then, there exist well-defined hyperparameters $\hat{\eta}$ and $\hat{\lambda}$, such that $\hat{A}$ is equivalent to $A^*$ with respect to $d(x, x')$, and the resulting $\hat{p}^A(y \mid x, x')$ parameterized by $\hat{A}$ and $\hat{\eta}$ is identical to the original $p^A(y \mid x, x')$ parameterized by $A^*$, $\kappa^*$ and $\eta$.*

*Remark* 1. After the simplification, $\gamma$ is dropped, $\eta$ and $\lambda$ are modified, but the regularization parameter $\mu$ remains the same, which means that the tradeoff between the supervised and unsupervised parts has not been affected.

Optimization (7) could be directly solved by the gradient projection method (Polyak, 1967), even though it is nonconvex. Nevertheless, we would like to pose it as an EM-like iterative scheme to access the derandomization by the initial solution, the stability for the gradient update, and the insensitivity to the step size, just to name a few of the gained algorithmic properties.

The EM-like algorithm runs as follows. In the beginning, we initialize a nonparametric probability $q(y \mid x_i, x_j)$, and then the M-Step and the E-Step get executed repeatedly until the stopping conditions are satisfied.

At the $t$-th E-Step, similarly to Graça et al. (2009), we have for each pair $(x_i, x_j) \in \mathcal{U}$ that

$$\min_q \ \operatorname{KL}(q \| p_{i,j}^A) + \mu \mathbb{E}_q[-\ln p_{i,j}^A(y)], \quad (9)$$

where KL is the Kullback-Leibler divergence, and $p_{i,j}^A$ is parameterized by the metric $A^{(t)}$ found at the last M-Step. Optimization (9) can be solved analytically.

---

[5]The new functions and parameters are denoted by $\hat{\cdot}$ within this theorem for the sake of clarity.

**Theorem 3.** *The solution to* (9) *is given by*

$$q(y \mid x_i, x_j) = \frac{p_{i,j}^A(y) \exp(\mu \ln p_{i,j}^A(y))}{\sum_{y'} p_{i,j}^A(y') \exp(\mu \ln p_{i,j}^A(y'))}. \quad (10)$$

On the other hand, at the $t$-th M-Step, we find new metric $A^{(t)}$ through the probability $q(y \mid x_i, x_j)$ which is generated in the last E-Step and only defined for $(x_i, x_j) \in \mathcal{U}$:

$$\max_A \; \mathcal{F}(A) = \sum_{\mathcal{S} \cup \mathcal{D}} \ln p_{i,j}^A(y_{i,j}) \\ + \mu \sum_{\mathcal{U}} \sum_y q(y \mid x_i, x_j) \ln p_{i,j}^A(y) - \lambda \operatorname{tr}(A). \quad (11)$$

It could be solved by the gradient projection method without worry about local maxima using the calculation of $\nabla \mathcal{F}$ given by

$$\nabla \mathcal{F}(A) = -\sum_{\mathcal{S} \cup \mathcal{D}} y_{i,j} \left(1 - p_{i,j}^A(y_{i,j})\right) x_{i,j} \\ - \mu \sum_{\mathcal{U}} \sum_y y q(y \mid x_i, x_j) \left(1 - p_{i,j}^A(y)\right) x_{i,j} - \lambda I_m,$$

where $x_{i,j} = (x_i - x_j)(x_i - x_j)^\top$, since the convexity of the feature function $f(x, x', y; A)$ with respect to $A$ implies the convexity of the objective $\mathcal{F}(A)$.

A remarkable property of $\mathcal{F}(A)$ is that its gradient is uniformly bounded, regardless of the scale of $A$, i.e., the magnitude of $\operatorname{tr}(A)$.

**Theorem 4.** *The objective $\mathcal{F}(A)$ is Lipschitz continuous, and the best Lipschitz constant $\operatorname{Lip}_{\|\cdot\|_F}(\mathcal{F})$ with respect to the Frobenius norm $\|\cdot\|_F$ satisfies*

$$\operatorname{Lip}_{\|\cdot\|_F}(\mathcal{F}) \leq (\#\mathcal{S} + \#\mathcal{D} + \mu \#\mathcal{U})(\operatorname{diam}(\mathcal{X}))^2 + \lambda m, \quad (12)$$

*where $\operatorname{diam}(\mathcal{X}) = \max_{x_i, x_j \in \mathcal{X}} \|x_i - x_j\|_2$ is the diameter of $\mathcal{X}$, and $\#$ measures the cardinality of a set.*

In our current implementation, the initial solution is $q(-1 \mid x_i, x_j) = 1$, which means that we treat all unlabeled pairs as dissimilar pairs. The overall asymptotic time complexity is $O(n^2 m + m^3)$ in which the stopping criteria of the M-Step and the whole EM-like iteration are ignored. Discussions about the computational complexity and the fast implementation can be found in Appendix D.

## 3. Discussions

In this section, we discuss the sparsity issues, namely, we can obtain the *posterior sparsity* (Graça et al., 2009) by entropy regularization and the *projection sparsity* (Ying et al., 2009) by trace-norm regularization.

By a 'sparse' posterior distribution, we mean that the uncertainty (i.e., the entropy or variance) is low. See Figure 1 as an example. Recall that supervised metric learning aims at a metric under which data in the same class are close and data from different classes are far apart. This results in the metric which ignores the horizontal feature and focuses on the vertical feature. However, the vertical feature is important, and taking care of the posterior sparsity would lead to a better metric as illustrated in (e) and (f). Therefore, we prefer taking the posterior sparsity into account in addition to the aforementioned goal, and then the risk of overfitting weakly labeled data can be significantly reduced.

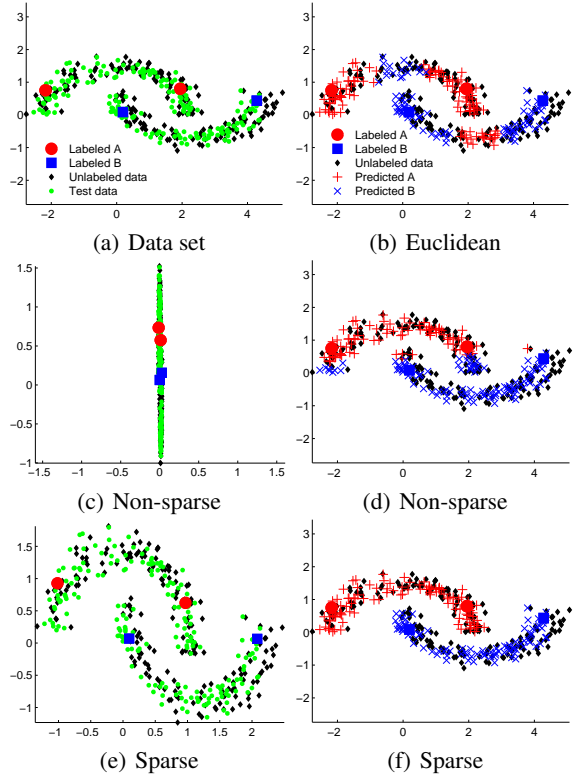

*Figure 1.* Sparse vs. non-sparse posterior distributions. Six weak labels are constructed according to four class labels. The left three panels show the original data and the projected data by metrics learned with/without the posterior sparsity. The right three panels exhibit one-nearest-neighbor classification results based on the Euclidean distance and two learned metrics.

We can rewrite $\mathcal{L}_2(A, \kappa)$ as a soft posterior regularization (PR) objective (Graça et al., 2009). Let the auxiliary feature function be $g(x, x', y) = -\ln p^A(y \mid x, x')$, then maximizing $\mathcal{L}_2(A, \kappa)$ is equivalent to

$$\max_{A, \kappa} \mathcal{L}_1(A, \kappa) - \mu \sum_{\mathcal{U}} \mathbb{E}_{p_{i,j}^A}[g(x_i, x_j, y)]. \quad (13)$$

On the other hand, according to optimization (7) of Graça et al. (2009), the soft PR objective should take a form as

$$\max_{A, \kappa} \; \mathcal{L}_1(A, \kappa) - \min_q \left( \operatorname{KL}(q \,\|\, p^A) + \mu \sum_{\mathcal{U}} \xi_{i,j} \right) \\ \text{s.t.} \; \mathbb{E}_q[g(x_i, x_j, y)] \leq \xi_{i,j}, \forall (x_i, x_j) \in \mathcal{U}, \quad (14)$$

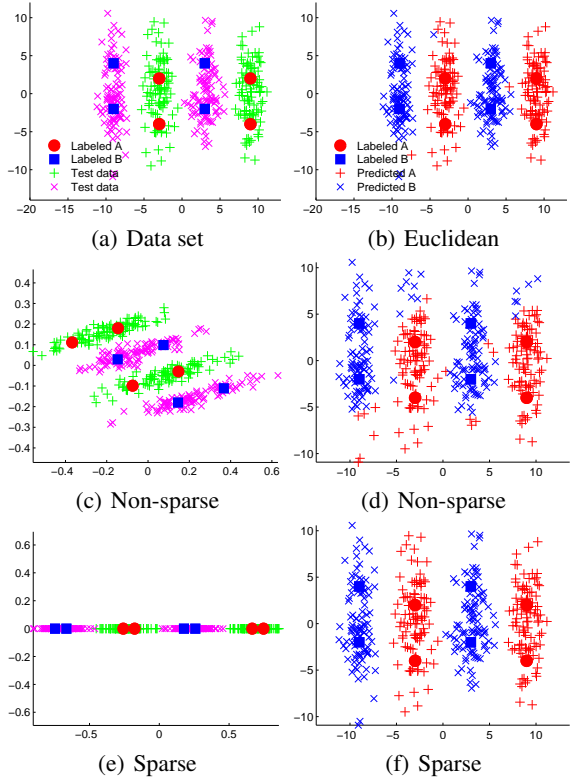

Figure 2. Sparse vs. non-sparse projections. The settings and the layout of panels are similar to Figure 1.

where $\xi_{i,j}$ are slack variables. Since $q$ is unconstrained, we can optimize it with respect to fixed $A$ and $\kappa$. It is easy to see that $q$ should be $p^A$ (restricted on $\mathcal{U}$), so the KL term is zero and the expectation term is the entropy, which implies the equivalence of optimizations (13) and (14).

Besides the above posterior sparsity, we also hope for the projection sparsity, which may guide the learned metric to better generalization performance. See Figure 2 as an example of its effectiveness, where the horizontal feature is informative and the vertical feature is useless.

The underlying technique is the mixed-norm regularization (Argyriou et al., 2006). Denote the $\ell_{(2,1)}$-norm of a symmetric matrix $M$ as $\|M\|_{(2,1)} = \sum_{k=1}^{m} (\sum_{k'=1}^{m} M_{k,k'}^2)^{1/2}$. Similarly to Ying et al. (2009), let $P \in \mathbb{R}^{m \times m}$ be a projection, and $W = P^\top P$ be the metric induced from $P$. Let the $i$-th column of $P$ and $W$ be $P_i$ and $W_i$. If $P_i$ is identically zero, the $i$-th component of $x$ has no contribution to $z = Px$. Since the column-wise sparsity of $W$ and $P$ are equivalent, we can penalize $\|W\|_{(2,1)}$ to reach the column-wise sparsity of $P$.

Nevertheless, this is feature selection rather than dimensionality reduction. Recall that the goal is to select a few most representative directions of input data which are not restricted to the coordinate axes. The solution is to pick an extra transformation $V \in \mathcal{O}^m$ to rotate $x$ before the projection where $\mathcal{O}^m$ is the set of orthonormal matrices of size $m$, and add $V$ to the optimization variables. Consequently, we penalize $\|W\|_{(2,1)}$, project $x$ to $z = PVx$, and since $A = (PV)^\top(PV) = V^\top WV$, we arrive at

$$\max_{A,\kappa,W,V} \mathcal{L}_2(A,\kappa) - \lambda \|W\|_{(2,1)} \quad (15)$$
$$\text{s.t.} \quad A = V^\top WV, W = W^\top, W \succeq 0, V \in \mathcal{O}^m.$$

The equivalence of optimizations (6) and (15) is guaranteed by Lemma 1 of Ying et al. (2009).

Moreover, there is another justification based on the *information maximization principle* (Gomes et al., 2010). Please see Appendix B.2 for details.

## 4. Related Works

Xing et al. (2002) initiated metric learning based on pairwise similarity/dissimilarity constraints by global distance metric learning (GDM). Several excellent metric learning methods have been developed in the last decade, including neighborhood component analysis (NCA; Goldberger et al., 2004), large margin nearest neighbor classification (LMNN; Weinberger et al., 2005), and information-theoretic metric learning (ITML; Davis et al., 2007).

Both ITML and SERAPH are information-theoretic, but the ideas and models are quite different. ITML defines a generative model $p^A(x) = \exp(-\frac{1}{2}\|x - \mu\|_A^2)/Z$, where $\mu$ is unknown mean value and $Z$ is a normalizing constant. Compared with GDM, ITML regularizes the KL-divergence between $p^{A_0}(x)$ and $p^A(x)$, and transforms this term to a Log-Det regularization. By specifying $A_0 = \frac{1}{n}I_m$, it becomes the maximum entropy estimation of $p^A(x)$. Thus, it prefers the metric close to the Euclidean distance. SERAPH also follows the maximum entropy principle, but the probabilistic model $p^A(y \mid x, x')$ is discriminative.

A probabilistic GDM was designed intuitively as a baseline in the experimental part of Yang et al. (2006). It is a special case of our supervised part. In fact, SERAPH is much more general. Please refer to Section 2.2 for details.

Subsequently, local distance metric learning (LDM; Yang et al., 2006) is the pioneer of semi-supervised metric learning, which assumes that the eigenvectors of $A$ are the principal components of training data. Hoi et al. (2008) combines manifold regularization to the min-max principle of GDM based on Belkin & Niyogi (2001), and Baghshah & Shouraki (2009) shows that Roweis & Saul (2000) is also useful for semi-supervised metric learning. Liu et al. (2010) brings the element-wise sparsity to Hoi et al. (2008).

The manifold extension described in Appendix C.3 can be attached to all metric learning methods, whereas our unsu-pervised part applies to probabilistic methods only. However, any probabilistic method with an explicit expression of the posterior distribution adopts two semi-supervised extensions, while deterministic methods such as LMNN cannot benefit from entropy regularization.

Table 1. Specification of benchmark data sets.

|  | #classes | #features ($m$) | #training ($n$) | #test | #class labels | $\mathbb{E}\#\mathcal{S}$ | $\mathbb{E}\#\mathcal{D}$ | $\#\mathcal{U}$ |
|---|---|---|---|---|---|---|---|---|
| iris | 3 | 4 | 100 | 38 | 10 | 15.10 | 29.90 | 4905 |
| wine | 3 | 13 | 100 | 78 | 10 | 13.98 | 31.02 | 4905 |
| ionosphere | 2 | 34 | 100 | 251 | 20 | 97.50 | 92.50 | 4760 |
| balance | 3 | 4 | 100 | 465 | 10 | 20.38 | 24.62 | 4905 |
| breast cancer | 2 | 30 | 100 | 469 | 10 | 23.54 | 21.46 | 4905 |
| diabetes | 2 | 8 | 100 | 668 | 10 | 23.02 | 21.98 | 4905 |
| USPS$_{1-5,20}$ | 5 | 64 | 100 | 2500 | 10 | 5 | 40 | 4905 |
| USPS$_{1-5,40}$ | 5 | 64 | 200 | 2500 | 20 | 30 | 160 | 19710 |
| USPS$_{1-10,20}$ | 10 | 64 | 200 | 2500 | 20 | 10 | 180 | 19710 |
| USPS$_{1-10,40}$ | 10 | 64 | 400 | 2500 | 40 | 60 | 720 | 79020 |
| MNIST$_{1,7}$ | 2 | 196 | 100 | 1000 | 4 | 2 | 4 | 4944 |
| MNIST$_{3,5,8}$ | 3 | 196 | 150 | 1500 | 9 | 9 | 27 | 11139 |

Due to limited space, we leave out sparse metric learning and robust metric learning. Instead, we recommend Huang et al. (2009) and Huang et al. (2010) for the latest reviews of sparse and robust metric learning respectively.

## 5. Experiments

### 5.1. Setup

We compared SERAPH with the Euclidean distance, four famous supervised and two representative semi-supervised metric learning methods[6]: global distance metric learning (GDM; Xing et al., 2002), neighborhood component analysis (NCA; Goldberger et al., 2004), large margin nearest neighbor classification (LMNN; Weinberger et al., 2005), information-theoretic metric learning (ITML; Davis et al., 2007), local distance metric learning (LDM; Yang et al., 2006), and manifold Fisher discriminant analysis (MFDA; Baghshah & Shouraki, 2009).

Table 1 describes the specification of the data sets used in our experiments. The top six data sets (i.e., iris, wine, ionosphere, balance, breast cancer, and diabetes) come from the *UCI machine learning repository*[7], while the *USPS* and *MNIST* come from the homepage of the late Sam Roweis[8]. Gray-scale images of handwritten digits are downsampled to $8 \times 8$ and $14 \times 14$ pixel resolution resulting in 64- and 196-dimensional vectors for USPS and MNIST. The sym-bol USPS$_{1-5,20}$ means 20 training data from each of the first 5 classes, USPS$_{1-10,40}$ means 40 training data from each of all 10 classes, MNIST$_{1,7}$ means digits 1 versus 7, and so forth. Note that in the last two tasks, the dimensionality of data is greater than the size of all training data.

---

[6]We downloaded the codes of all baseline methods from the official websites provided by the original authors except MFDA.
[7]http://archive.ics.uci.edu/ml/.
[8]http://cs.nyu.edu/~roweis/data.html.

In our experiments, all methods were repeatedly run on 50 random samplings. For each random sampling, class labels of the first few data were revealed, and the sets $\mathcal{S}$ and $\mathcal{D}$ were constructed according to these revealed class labels. The sizes of $\mathcal{S}$, $\mathcal{D}$ and $\mathcal{U}$ were fixed for all samplings of USPS and MNIST but random for the samplings of UCI data sets. We measured the performance of the one-nearest-neighbor classifiers based on the learned metrics as well as the computation time for learning the metrics.

Four settings of SERAPH were included in our experiments (except on two artificial data sets): SERAPH$_{none}$ stands for $\mu = \lambda = 0$, SERAPH$_{post}$ for $\mu = \frac{\#(\mathcal{S} \cup \mathcal{D})}{\#\mathcal{U}}$ and $\lambda = 0$, SERAPH$_{proj}$ for $\mu = 0$ and $\lambda = 1$, and SERAPH$_{hyper}$ for $\mu = \frac{\#(\mathcal{S} \cup \mathcal{D})}{\#\mathcal{U}}$ and $\lambda = 1$. We fixed $\eta = 1$ for simplicity.

There was no cross-validation for each random sampling, otherwise the learned metrics would be highly dependent upon the final classifier, and also because of the large variance of the classification performance given the limited supervised information. The hyperparameters of other methods, e.g., the number of reduced dimensions, the number of nearest neighbors, and the percentage of principal components, were selected as the best value based on another 10 random samplings if default values or heuristics were not provided by the original authors.

### 5.2. Results

Figures 1 and 2 had previously displayed the visually comprehensive results of the sparsity regularization on two artificial data sets respectively. Subfigures (c) and (d) in both

*Table 2.* Means with standard errors of the nearest-neighbor misclassification rate (in %) on UCI, USPS and MNIST benchmarks. For each data set, the best method and comparable ones based on the unpaired $t$-test at the significance level 5% are highlighted in boldface.

| | iris | wine | ionosphere | balance | breast cancer | diabetes |
|---|---|---|---|---|---|---|
| EUCLIDEAN | $9.58 \pm 0.73$ | $12.93 \pm 0.83$ | $23.60 \pm 0.89$ | $27.15 \pm 0.75$ | $14.11 \pm 1.07$ | $32.94 \pm 0.65$ |
| GDM | $8.95 \pm 0.71$ | $11.52 \pm 0.77$ | $\mathbf{20.82 \pm 0.82}$ | $22.89 \pm 1.08$ | $11.86 \pm 0.83$ | $\mathbf{30.73 \pm 0.59}$ |
| NCA | $10.32 \pm 0.83$ | $15.03 \pm 1.12$ | $26.68 \pm 0.82$ | $32.97 \pm 1.31$ | $14.63 \pm 1.09$ | $32.95 \pm 0.65$ |
| LMNN | $9.81 \pm 0.79$ | $14.83 \pm 0.97$ | $22.25 \pm 0.75$ | $24.00 \pm 1.34$ | $13.86 \pm 0.84$ | $32.02 \pm 0.60$ |
| ITML | $\mathbf{5.57 \pm 0.53}$ | $\mathbf{8.22 \pm 0.66}$ | $20.35 \pm 0.64$ | $22.04 \pm 0.80$ | $\mathbf{9.60 \pm 0.49}$ | $31.21 \pm 0.73$ |
| LDM | $7.27 \pm 0.72$ | $17.21 \pm 1.41$ | $24.54 \pm 0.92$ | $\mathbf{21.22 \pm 0.93}$ | $14.85 \pm 0.92$ | $34.33 \pm 0.60$ |
| MFDA | $6.58 \pm 0.54$ | $11.55 \pm 1.03$ | $23.66 \pm 0.91$ | $23.61 \pm 1.00$ | $11.21 \pm 0.80$ | $31.64 \pm 0.62$ |
| SERAPH$_{none}$ | $6.21 \pm 0.48$ | $\mathbf{8.13 \pm 0.58}$ | $19.70 \pm 0.43$ | $20.25 \pm 0.64$ | $11.39 \pm 0.49$ | $29.86 \pm 0.61$ |
| SERAPH$_{post}$ | $\mathbf{4.79 \pm 0.37}$ | $\mathbf{7.46 \pm 0.51}$ | $19.64 \pm 0.45$ | $19.98 \pm 0.67$ | $11.33 \pm 0.50$ | $29.87 \pm 0.57$ |
| SERAPH$_{proj}$ | $\mathbf{5.79 \pm 0.54}$ | $\mathbf{7.39 \pm 0.50}$ | $19.53 \pm 0.46$ | $20.94 \pm 0.64$ | $\mathbf{9.61 \pm 0.49}$ | $30.43 \pm 0.65$ |
| SERAPH$_{hyper}$ | $\mathbf{5.31 \pm 0.43}$ | $\mathbf{7.38 \pm 0.49}$ | $19.33 \pm 0.42$ | $20.15 \pm 0.63$ | $10.04 \pm 0.52$ | $30.02 \pm 0.63$ |
| | USPS$_{1-5,20}$ | USPS$_{1-5,40}$ | USPS$_{1-10,20}$ | USPS$_{1-10,40}$ | MNIST$_{1,7}$ | MNIST$_{3,5,8}$ |
| EUCLIDEAN | $36.63 \pm 0.80$ | $28.43 \pm 0.60$ | $49.17 \pm 0.50$ | $39.30 \pm 0.39$ | $10.42 \pm 0.67$ | $\mathbf{37.30 \pm 0.81}$ |
| GDM | $37.62 \pm 0.77$ | - | - | - | - | - |
| NCA | $37.55 \pm 0.84$ | $28.39 \pm 0.60$ | $57.01 \pm 0.82$ | $49.21 \pm 0.66$ | $10.42 \pm 0.67$ | $\mathbf{37.75 \pm 0.92}$ |
| LMNN | $36.43 \pm 0.78$ | $28.93 \pm 0.61$ | $48.12 \pm 0.57$ | $43.68 \pm 0.58$ | $9.99 \pm 0.71$ | $\mathbf{36.49 \pm 0.82}$ |
| ITML | $35.86 \pm 0.74$ | $27.40 \pm 0.65$ | $47.40 \pm 0.60$ | $39.44 \pm 0.57$ | $9.94 \pm 0.69$ | $40.83 \pm 0.93$ |
| LDM | $47.19 \pm 1.51$ | $32.52 \pm 0.85$ | $59.13 \pm 0.73$ | $43.18 \pm 0.53$ | $14.54 \pm 1.41$ | $45.53 \pm 1.16$ |
| MFDA | $42.52 \pm 0.82$ | $28.82 \pm 0.62$ | $52.13 \pm 0.59$ | $37.78 \pm 0.50$ | $9.35 \pm 0.72$ | $42.39 \pm 0.92$ |
| SERAPH$_{none}$ | $36.08 \pm 0.75$ | $27.41 \pm 0.60$ | $47.29 \pm 0.58$ | $38.36 \pm 0.55$ | $9.97 \pm 0.71$ | $\mathbf{36.44 \pm 0.84}$ |
| SERAPH$_{post}$ | $35.79 \pm 0.75$ | $27.37 \pm 0.60$ | $47.12 \pm 0.58$ | $38.20 \pm 0.55$ | $10.98 \pm 0.79$ | $\mathbf{36.45 \pm 0.84}$ |
| SERAPH$_{proj}$ | $36.01 \pm 0.75$ | $\mathbf{26.17 \pm 0.57}$ | $47.42 \pm 0.62$ | $35.42 \pm 0.54$ | $9.28 \pm 0.72$ | $\mathbf{36.55 \pm 0.80}$ |
| SERAPH$_{hyper}$ | $\mathbf{32.79 \pm 0.77}$ | $\mathbf{25.26 \pm 0.56}$ | $\mathbf{44.89 \pm 0.58}$ | $\mathbf{33.41 \pm 0.47}$ | $\mathbf{7.61 \pm 0.57}$ | $35.71 \pm 0.84$ |

figures were obtained by GDM, while (e) and (f) were generated by SERAPH with $\mu = 10 \cdot \frac{\#(\mathcal{S} \cup \mathcal{D})}{\#\mathcal{U}}$, $\lambda = 0$ in Figure 1 and $\mu = 0, \lambda = 300$ in Figure 2. We can see from Figures 1 and 2 that SERAPH improved supervised global metric learning dramatically by the sparsity regularization.

The experimental results of the one-nearest-neighbor classification are reported in Table 2 (GDM was sometimes very slow and excluded from the comparison). SERAPH is fairly promising, especially with the hyper-sparsity ($\mu = \frac{\#(\mathcal{S} \cup \mathcal{D})}{\#\mathcal{U}}$ and $\lambda = 1$). It was best or tie over all tasks, and often statistically significantly better than others on UCI data sets except ITML. It was better than all other methods statistically significantly on USPS, and SERAPH$_{hyper}$ outperformed both SERAPH$_{post}$ and SERAPH$_{proj}$. Moreover, it improved the accuracy even on the ill-posed MNIST tasks, though the improvement was insignificant on MNIST$_{3,5,8}$. In a word, SERAPH can reduce the risk of overfitting weakly labeled data with the help of unlabeled data, and hence our sparsity regularization would be reasonable and practical.

In vivid contrast with SERAPH that exhibited nice generalization capability, supervised methods might learn a metric even worse than the Euclidean distance due to overfitting problems, especially NCA that optimized the leave-one-out performance based on such limited label information. The powerful LMNN did not behave satisfyingly, since it was hardly fulfilled to find a lot of neighbors belonging to the same class within labeled data. ITML was the second best method though it can only access weakly labeled data, but it became less useful for difficult tasks. On the other hand, we observed that LDM might fail when the principal components of training data were not close to the eigenvectors of the target matrix, and MFDA might fail if the amount of training data cannot recover the underlying manifold well.

An observation is that the global metric learning often outperformed the local one, if the supervised information was insufficient. This phenomenon indicates that the local metric learning tends to fit the local neighborhood information exceedingly and then suffers from overfitting problems.

Finally, we report in Figure 3 the computation time of each algorithm on each task (excluding GDM). Generally speaking, SERAPH was the second fastest method, and the fastest MFDA involves only some matrix multiplication and a single eigen-decomposition. Improvements may be expected if we program in Matlab with C/C++.

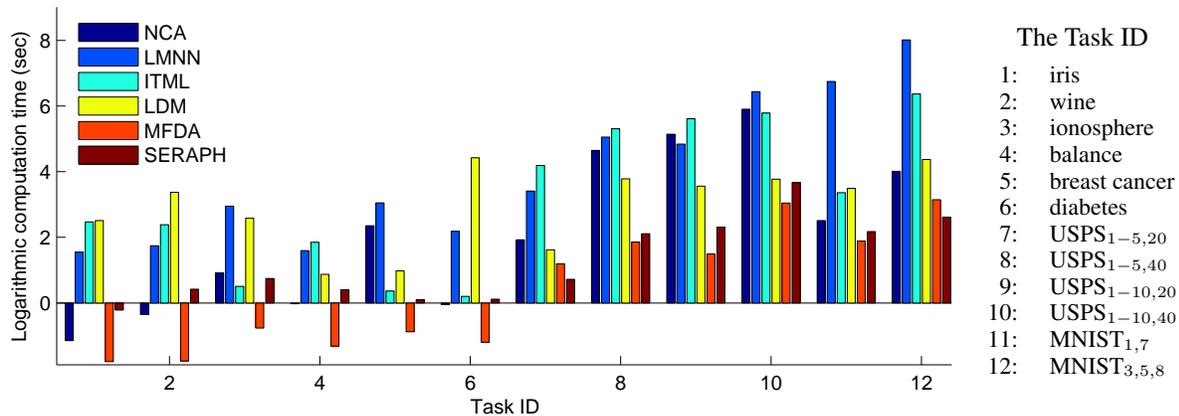

Figure 3. Computation time (per run) of different metric learning algorithms.

## 6. Conclusions

In this paper, we proposed an information-theoretic semi-supervised metric learning approach SERAPH as an alternative to the manifold-based methods. The generalized maximum entropy estimation for supervised metric learning was our foundation. Then a semi-supervised extension that can achieve the posterior sparsity was obtained via entropy regularization. Moreover, we enforced a trace-norm regularization that can reach the projection sparsity. The resulting optimization was solved by an EM-like scheme with several nice algorithmic properties, and the learned metric had high discriminability even under a noisy environment.

Experiments on benchmark data sets showed that SERAPH often outperformed state-of-the-art fully-/semi-supervised metric learning methods given only limited supervised information. A final note is that in our experiments the posterior and projection sparsity were demonstrated to be very helpful for high-dimensional data *if and only if* they were combined with each other, i.e., integrated into the hyper-sparsity. An in-depth study of this interaction is left as our future work.

## Acknowledgments

The authors would like to thank anonymous reviewers for helpful comments. GN is supported by the MEXT scholarship No.103250, MY is supported by the JST PRESTO program, and MS is supported by the FIRST program.